\documentclass{article}

\usepackage[dblblindworkshop, final]{neurips_2026}
\workshoptitle{Continual Learning for Enterprise AI Agents (CLEA)}

\usepackage[utf8]{inputenc}
\usepackage[T1]{fontenc}
\usepackage{url}
\usepackage{booktabs}
\usepackage{amsfonts}
\usepackage{nicefrac}
\usepackage{microtype}
\usepackage{xcolor}
\usepackage{amsmath}
\usepackage{graphicx}
\usepackage{multirow}
\usepackage{natbib}
\usepackage{tikz}
\usetikzlibrary{shapes.geometric, arrows.meta, positioning, fit, backgrounds}

\title{Continuous Process-Level Evaluation for\\
       Evolving Enterprise AI Agent Skills}

\author{%
  Ngoc Phuoc An Vo \\
  IBM\\
  \texttt{ngoc.phuoc.an.vo@ibm.com} \\
  \and
  Aarya Doshi\thanks{Work done during a 2026 summer internship at IBM Research.} \\
  Georgia Institute of Technology \\
  \texttt{aaryadoshi@gmail.com} \\
  \and
  Vadim Sheinin \\
  IBM \\
  \texttt{vadims@us.ibm.com} \\
}

\begin{document}

\maketitle

\begin{abstract}
Enterprise AI agent skills are not static artifacts: they are continually
revised as tool APIs change, LLM versions update, and skill specifications
are refined in response to operational feedback.
Yet the standard practice for validating each revision is to check only
final-output accuracy---an approach that systematically misses process-level
behavioral drift introduced during evolution.
We present a \emph{continuous evaluation framework} for enterprise agentic
skills that combines outcome-level and process-level quality checks,
applied to two \emph{Business Value Determination}~(BVD) skill variants
within an enterprise \emph{Value Aware Resiliency}~(VAR) system.
The framework independently computes ground truth per run, instantiates
template test cases as persistent regression tests, and assesses tool
selection, argument correctness, execution ordering, and database integrity
using programmatic checks augmented by a narrowly scoped LLM judge when
exact matching would be brittle.
We report a fully automatic evaluation over 240~trials spanning
two related but structurally distinct skills (Revenue and Productivity
apportionment), two specification variants (SKILL.md and Skill.txt), two agent harnesses
(Claude~Code and Codex), and three models
(GPT-5.6-Sol, Claude~Opus~4.8, Claude~Sonnet~4.6).
Across the 240 automatic trials, 175 passed all applicable final numerical
checks; among them, 162 (92.6\%; Wilson 95\% CI: 87.7--95.6\%) still had
at least one additional evaluator-detected deviation. Under a broader
seven-check final-state definition, 151/164 passing runs (92.1\%; 95\% CI:
86.9--95.3\%) still violated a trajectory check. Dependency attribution
descriptively compressed a mean 6.34 failed checks per run to 2.65 roots. Specification-variant
sensitivity also varies with model and
harness: unadjusted bootstrap interaction intervals exclude zero for all three
Revenue comparisons and for GPT on Productivity, but not for the two
remaining Productivity comparisons.
Template test cases with runtime-resolved placeholders provide reusable
regression coverage across the evaluated specifications, models, and harnesses;
longitudinal validation under actual API evolution remains future work.
\end{abstract}

\section{Introduction}

Enterprise AI agent skills are continuously evolving systems.
The skill that handles IT infrastructure cost apportionment today will be
revised tomorrow when the observability API version changes, again next
month when a new formula is agreed on with the finance team, and again
when the underlying LLM is upgraded to the next model version.
Each of these changes is an opportunity to introduce behavioral
regressions---deviations from the intended process that the skill was
designed to follow.

Existing evaluation practice addresses this challenge poorly.
Standard evaluation checks whether the final output---a report, a database
row, a dollar figure---matches the expected value.
But an agent skill can produce a correct-looking final output while taking
the wrong steps to get there: calling APIs with incorrect scope,
skipping a required database readback, writing intermediate results to
the wrong table.
These \emph{process deviations} are invisible to outcome-only checks yet
represent genuine behavioral drift that can corrupt downstream agent steps,
violate data integrity constraints, or produce correct-by-coincidence outputs
that will fail when inputs change slightly.
In a continual-adaptation setting---where skills are revised repeatedly
in response to operational feedback---these failures compound across
revisions, making process-level evaluation an essential component of any
responsible deployment loop.

This paper studies the evaluation layer needed to support continual agent
evolution safely, rather than proposing a continual-learning algorithm.
We make the case that \emph{continuous process-level evaluation} is an
appropriate quality gate for enterprise agent skills as they evolve.
We demonstrate the principle on the Business Value Determination~(BVD) skill
in an enterprise Value Aware Resiliency~(VAR) system.
Even when BVD produces the expected final allocation, its trace can violate
required scope, persistence, or ordering constraints. A portfolio-total-only
outcome check cannot detect these deviations, although finer-grained outcome
and process checks can.

We contribute:
\begin{itemize}
  \item An \textbf{environment-grounded evaluation framework} that checks
    tool selection, arguments, ordering, and database integrity against an
    oracle computed independently of the evaluated trajectory, augmented by
    a narrowly scoped LLM judge for semantically flexible arguments.
  \item A \textbf{parameterized regression contract} whose runtime-resolved
    placeholders materialize expected scope, multiplicity, and values rather
    than hardcoding them, with one registered suite reused across all tested
    configurations (104~cases for Revenue; 80 for Productivity).
  \item Empirical evidence---240~automatic trials spanning two specification
    variants, two harnesses, and three models---that
    162 of 175 trials that pass all applicable final numerical checks still
    have at least one additional evaluator-detected deviation; even after
    broadening the baseline to seven final-state checks per skill, 151 of 164
    passing runs violate a trajectory check.
  \item Sensitivity analyses and root-family decomposition showing that the
    outcome/process gap persists after excluding external-error runs and spans
    data consistency, missing phases, allowlist, call-count, and scope checks.
\end{itemize}

\section{Enterprise Agent Skills as Evolving Systems}

Enterprise AI agent skills differ from research benchmarks in a critical
way: they are not evaluated once and retired.
They are revised on an ongoing basis in response to API changes,
specification refinements, new tool capabilities, and LLM model
upgrades~\citep{bogavelli2025agentarch}.
We characterize four distinct \emph{evolution axes} for enterprise skills:

\paragraph{Tool API evolution.}
Observability platforms, databases, and business systems update their
APIs continuously.
An agent skill that correctly calls a tool with today's argument schema
may break silently when the tool's parameter names or scoping semantics
change in the next platform release.

\paragraph{Skill specification revision.}
As operations teams learn from skill behavior in production, they update
the skill's specification document to codify learned behaviors, close
ambiguities, or change business logic.
Each revision risks changing the agent's planning strategy in ways not
captured by the outcome metric.

\paragraph{LLM version change.}
When the underlying model is updated---even within a model family---the
agent's default planning choices for ambiguous steps can shift.
A step that one model version resolved conservatively might be resolved
aggressively by the next, changing which tools are called and in what order.

\paragraph{Harness change.}
Enterprise deployments may migrate between agent harnesses
(e.g., Claude~Code to Codex or vice versa) for cost, latency, or
integration reasons.
As we show, harnesses with similar outcome accuracy exhibit distinct
process-level deviation patterns associated with different execution
characteristics, and the same specification variant can be robust under one
harness while causing significant accuracy drops under another.

Each evolution axis introduces risk that outcome-only evaluation
cannot detect.
A quality gate that checks only the final output approves skill revisions
that are behaviorally regressed at the process level.
We propose that \emph{process-level evaluation against persistent template
test cases} is the correct quality gate for this setting.

\section{Related Work}

\paragraph{Behavioral drift and stability in agentic systems.}
Agents produce unstable outputs across repeated runs on identical inputs;
aggregate task-completion metrics can mask tool-use, policy, verification,
and memory-recall failures~\citep{bogavelli2025agentarch,akshathala2026beyond}.
In the multi-agent traces analyzed by~\citet{cemri2026multi}, system-design
and task-verification categories account for 63.1\% of annotated failures
(41.8\% and 21.3\%, respectively). This distribution identifies substantial
system-level improvement opportunities, but does not rule out limitations of
the underlying models.

\paragraph{Evaluation methodology for agents.}
\citet{flynt2026groundeval} proposes GroundEval, a deterministic
alternative to LLM-as-judge for stateful tasks. GroundEval scores an observed
trace and final answer against a human-reviewed, machine-checkable state
contract covering evidence, time, access, and required searches, rather than
against a canonical reference trajectory. Evaluation methods provide
complementary signals: deterministic checks offer reproducibility, LLM judges
accommodate semantic variation, and human review supports calibration and
oversight. Accordingly, \citet{gritta2026process} advocate combining outcome
and process evaluation, whereas \citet{zheng2023judging} advocate combining
capability-based and preference-based benchmarks.

\paragraph{LLM-as-judge limitations.}
Prior studies show that LLM-based evaluation can be sensitive to factors
unrelated to response quality, including candidate-response order,
verbosity, and model identity~\citep{wang2024large,zheng2023judging}.
Deterministic rules can instead miss semantically expressed failures;
for example, AgentEval reports 0.58 failure-detection recall for 47
hand-crafted rules against human annotations~\citep{guo2026agenteval}.
Our framework combines rule-based structural checks with LLM validation of
selected semantically flexible arguments and optional failure-only rescue
audits. Two authors manually reviewed 60 randomly sampled LLM-judged cases;
57 verdicts agreed with their assessment (95\%). This targeted audit
reduces, but does not eliminate, judge-validity risk.

\paragraph{Reference integrity.}
In LLM-generated tests for HumanEval, 88.89\% of erroneous test cases
studied by~\citet{huang2025nexus} arose from incorrect generated output
oracles rather than invalid test inputs; the corresponding oracle-error
share ranged from 79.71\% to 100\% across their seven benchmarks.
Our framework computes ground truth by calling the same real APIs
the skill uses, \emph{independently} of the agent execution trace,
avoiding dependence on an LLM-generated oracle while retaining the
separate risk that live state can change between execution and evaluation.

\paragraph{Process versus outcome evaluation.}
Grading only the final output can reward a correct-looking result
reached through a noncompliant or unsupported process. Process evaluation
has consequently been advocated for critical agent applications, while
process supervision has outperformed outcome-only supervision in mathematical
reasoning~\citep{gritta2026process,lightman2024verify}.
Trajectory-level diagnosis of agent failures from execution traces is
demonstrated by~\citet{barke2026agentrx}.

\paragraph{CI/CD-style regression evaluation.}
Automated regression testing and continuous monitoring in production ML
provide a software-engineering precedent for continuous skill quality
assurance~\citep{breck2017mltest}.
\citet{guo2026agenteval} apply this idea to agentic tasks through
AgentEval, which uses a dependency-modeled DAG to chain step-level checks
and improve root-cause accuracy, making it closely related to our framework.
We extend this direction with two additions absent from AgentEval:
(i)~\emph{runtime-resolved template test cases} designed to be reused as
skill specifications, tool APIs, and input portfolios evolve; and
(ii)~\emph{live-API ground truth} computed independently of the agent
execution, designed to avoid dependence on self-generated oracles---a failure mode illustrated by high generated-oracle error shares in automated test generation~\citep{huang2025nexus}.
Unlike AgentEval, whose evaluation DAG is instantiated around a specific
workflow execution, our regression artifact resolves expected values from the
current enterprise environment at evaluation time. Our experiments establish
reuse across configurations, not yet longitudinal persistence under API change.

\section{System: VAR and BVD}

\paragraph{Value Aware Resiliency (VAR).}
VAR is an enterprise resiliency agent system that measures application
resilience against business-defined SLOs.
It consists of a pipeline of LLM-powered agentic skills: Business Value
Determination (BVD), Assess Attribution, Application Optimizer, Diagnosis,
Recommendation Generation, Recommendation Impact Estimator, and
Recommendation Optimizer.
Skills are implemented as Claude~Code or Codex sessions directed by
a skill specification document and equipped with tools from two MCP
servers: the \emph{VAR MCP Server} (23~tools; stateless; fetches
utilization data from Instana, Kubecost, and ServiceNow) and the
\emph{VAR Data Access MCP Server} (18~tools; manages persistent
session state, result storage, and apportionment tables in SQLite).

\paragraph{Business Value Determination.}
BVD is Step~0 of VAR.
Given an application portfolio, a date range, and an annual revenue figure,
BVD queries observability APIs to retrieve per-application CPU, memory,
and call-count utilization, then allocates business value proportionally
(default weights: CPU~40\%, memory~40\%, calls~20\%).
Outputs are written to a persistent database read by all downstream skills.
Errors in BVD therefore corrupt the entire remaining pipeline.
We study two related but operationally different skills.
\textbf{Revenue Apportionment} allocates prorated portfolio revenue across
nine applications from CPU, memory, and call volume. It stages the complete
monitoring response, reads it back, computes allocation, and writes separate
host- and application-level revenue tables; its suite contains 104~checks.
\textbf{Productivity Apportionment} additionally retrieves Kubernetes costs
from Kubecost and EC2 unblended cost from Cloudability, requires EC2 resource
identifiers, splits host cost among applications, computes revenue-to-cost
productivity with explicit NULL semantics, and writes a different application
table; its suite contains 80~checks. The skills therefore share resource-based
revenue allocation but not the same tool set, persistence contract, or
workflow depth.

\paragraph{Specification variants.}
MD and TXT denote paired instruction artifacts, not a controlled file-format
manipulation. For Revenue, TXT is a substantial condensation (904 versus
1{,}379 words): it removes the tool-reference table and many exact output and
warning requirements while retaining staging, readback, allocation, and
persistence. For Productivity, TXT is only modestly shorter (1{,}318 versus
1{,}405 words) and adds operational content absent from MD, including an
explicit duration formula, the $[0.01,1]$ normalization range, exclusion of
non-portfolio applications, a combined EC2 cost query, and CSV export.
MD--TXT differences therefore estimate sensitivity to the complete revision
(syntax, length, and instruction content together), not a causal effect of
formatting alone. Both variants delegate deterministic arithmetic to tools.

\section{Continuous Evaluation Framework}

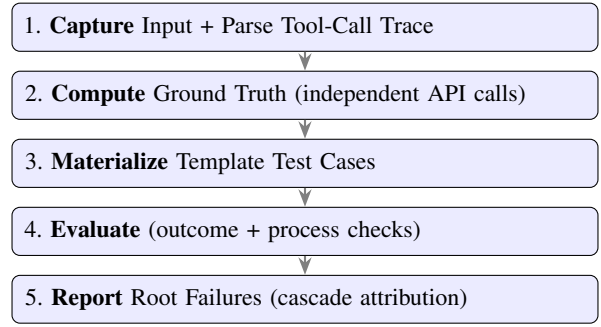
\begin{figure}[ht]
\centering
\begin{tikzpicture}[
  box/.style={draw, rounded corners=3pt, fill=blue!8,
              text width=7.4cm, minimum height=0.58cm,
              font=\small, inner sep=5pt},
  annot/.style={font=\footnotesize, text=gray, align=left,
                text width=5.0cm},
  arr/.style={-Stealth, thick, gray}
]
  \node[box] (s1) {1.~\textbf{Capture} Input + Parse Tool-Call Trace};
  \node[box, below=0.25cm of s1] (s2) {2.~\textbf{Compute} Ground Truth~(independent API calls)};
  \node[box, below=0.25cm of s2] (s3) {3.~\textbf{Materialize} Template Test Cases};
  \node[box, below=0.25cm of s3] (s4) {4.~\textbf{Evaluate}~(outcome + process checks)};
  \node[box, below=0.25cm of s4] (s5) {5.~\textbf{Report} Root Failures~(cascade attribution)};

  \draw[arr] (s1.south) -- (s2.north);
  \draw[arr] (s2.south) -- (s3.north);
  \draw[arr] (s3.south) -- (s4.north);
  \draw[arr] (s4.south) -- (s5.north);
\end{tikzpicture}
\caption{Continuous evaluation pipeline. Stages 1 and 5 are new for each
run; Stages 2--4 reuse a template suite within each skill across the evaluated
specification, model, and harness configurations.}
\label{fig:pipeline}
\end{figure}

The evaluation framework operates as a five-stage pipeline,
independent of the agent session under evaluation
(Figure~\ref{fig:pipeline}).

\subsection{Stage 1: Capture Input and Parse Tool-Call Trace}

A single line appended to the skill specification's first step directs the
agent to record all parsed input parameters (portfolio, date range, revenue)
to a structured log.
The agent's full tool-call trace---tool name, arguments, return value, and
conversation turn index---is captured from the harness execution log.

\subsection{Stage 2: Compute Ground Truth}

A standalone Python script calls the same real APIs the skill uses
via an independent execution path, and computes expected per-application
apportionment values.
This independence is critical: on HumanEval, 88.89\% of the erroneous
LLM-generated test cases analyzed by~\citet{huang2025nexus} were caused by
incorrect output oracles rather than invalid test inputs.
Live calls make the reference responsive to current API state, but execution
and evaluation can observe different snapshots; we treat this as a validity
threat rather than evidence of correctness under API evolution.

\subsection{Stage 3: Materialize Template Test Cases}

Template test cases describe intended skill behavior using
\emph{runtime-resolved placeholders} rather than hardcoded values.
A placeholder such as \texttt{\$k8s\_host\_date\_calls} resolves at
evaluation time from the ground truth computed in Stage~2.
This design makes templates input-parameterized: the same template file can
apply to different date ranges, portfolios, and revenue figures, provided the
intended tool semantics have not changed.
Dependency links between test cases define which failures are root causes
and which are downstream effects, enabling cascade root-cause attribution
in Stage~5.
For example, the following excerpt declares a conditional tool-call check;
the expected calls are filled from the run-specific oracle rather than copied
into the suite:
\begin{verbatim}
{"id": "k8s_cpu_mem",
 "condition": "$has_k8s",
 "check": {"type": "tool_call"},
 "tool": "...calculate_k8s_cpu_memory_usage",
 "scope_arg": "cluster_name",
 "expected_calls": "$k8s_host_date_calls"},
{"id": "compute_apps_match_stored_data",
 "depends_on": "resource_util_store_args_match",
 "check": {"type": "bash", ...}}
\end{verbatim}
Materialization expands this compact declaration into presence, argument,
scope, execution, redundancy, and call-count checks; downstream checks can
reference the expanded IDs through \texttt{depends\_on}.

\subsection{Stage 4: Evaluate}

\paragraph{Outcome and final-state checks.}
We distinguish three evaluation layers. Level~1 contains the pre-specified
final numerical outcomes: application revenue for Revenue, and application
revenue, total cost, and productivity for Productivity. Level~2 broadens this
to seven final-state checks per skill: the Level~1 values plus other stored
values and normalization sums, and required response structure. Level~3
contains the remaining trajectory checks described below. This separation
tests whether trajectory evaluation adds detection even after a richer final
state, rather than only after a deliberately narrow numerical baseline.

\paragraph{Process checks} cover four dimensions:
\begin{itemize}
  \item \textbf{Tool selection}: required tools called; no irrelevant
    tools invoked.
  \item \textbf{Tool arguments}: date ranges, filters, and numeric
    parameters checked with typed constraints.
    Candidate-call selection and structural checks are rule based. The LLM
    judge (\texttt{claude-sonnet-4-6}) is invoked only for designated checks
    where exact matching is too brittle, such as semantically equivalent SQL
    or alternative argument representations; its requested numeric
    comparisons are executed through a calculator tool. Selected failed
    programmatic checks may also receive a failure-only judge-rescue audit.
    It therefore supplements rather than replaces the programmatic suite and
    does not discover candidate calls. Two authors manually reviewed 60
    randomly sampled LLM-judged cases; 57 verdicts agreed with the authors'
    assessment (95\%).
  \item \textbf{Tool ordering}: required phase sequence enforced by
    comparing conversation turn indices (fetch $\to$ stage $\to$
    readback $\to$ allocate $\to$ write).
  \item \textbf{Argument scope}: per-cluster and per-host tool calls
    checked for under-coverage (missing calls) and over-coverage
    (calls with wrong scope values).
\end{itemize}

\paragraph{Why process checks matter for evolution.}
When a skill specification is condensed or a tool API changes, the model's
behavior at ambiguous steps shifts.
Outcome checks approve the new behavior if the final number is still right.
Process checks catch the behavioral shift itself---irrespective of whether
it accidentally produces the correct final output---and surface it as a
regression for human review before deployment.

\subsection{Stage 5: Report Root Failures with Cascade Attribution}

When a check fails, the framework traverses the authored dependency graph to
distinguish designated \emph{root failures} from \emph{cascaded failures}.
A cascaded failure is attributed to its root cause, preventing developers
from spending time on each downstream symptom independently.
The report lists roots with cascaded descendants to reduce the initial review
set; these graph-derived roots are descriptive, not validated causal explanations.

\section{Experimental Setup}

\paragraph{Skills and test suites.}
We evaluate two related BVD-family skills with independently computed ground truth:
\textbf{Revenue Apportionment} (104~template test cases) and
\textbf{Productivity Apportionment} (80~template test cases).
They share utilization retrieval and revenue allocation, but Productivity
adds cost retrieval, cost allocation, NULL handling, and a different
persistence contract (\S4).

\paragraph{Data.}
A real enterprise portfolio of nine applications across three infrastructure
platforms (two Kubernetes clusters, one EC2 host) and annual revenue
\$1{,}200{,}000. Revenue trials use 2026-05-01 through 2026-05-15;
Productivity trials use 2026-07-01 through 2026-07-15. Utilization is retrieved
live from Instana, while Productivity additionally queries Kubecost and
Cloudability. Comparisons are therefore made within skill; the study does not
interpret cross-skill differences as effects under a shared data snapshot.

\paragraph{Specification variants.}
Following \S4, we test \textbf{MD} (SKILL.md) and \textbf{TXT} (Skill.txt)
with computation tools enabled. Because each pair differs in presentation
and instructions, ``variant'' is the experimental factor.

\paragraph{Harnesses and models.}
We evaluate under two agent harnesses: \textbf{Claude~Code} (Anthropic CLI)
and \textbf{Codex} (OpenAI CLI).
Each harness is driven with three models: \textbf{GPT-5.6-Sol} (Azure),
\textbf{Claude~Opus~4.8} (AWS), and \textbf{Claude~Sonnet~4.6}.
This yields a $2\,{\times}\,2\,{\times}\,2\,{\times}\,3$ evaluation matrix
of skills, specification variants, harnesses, and models.

\paragraph{Trials.}
We run 10~trials per cell, all graded by the
evaluation framework against the corresponding template test suite.
The reported suite score is the fraction of all instantiated checks passed;
it aggregates outcome and process checks and should not be interpreted as
the probability that an entire run is correct.
One archived Productivity Codex--Sonnet TXT result contained six legacy
Cloudability checks not present in the registered 80-check suite. We removed
those six checks before aggregation, yielding a common suite within that skill.
Total automatic trials:
$2\,{\times}\,2\,{\times}\,2\,{\times}\,3\,{\times}\,10 = 240$.

\paragraph{Statistical analysis.}
We report trial-level means and standard deviations. To assess whether the
specification-variant effect changes across harnesses, we additionally compute the
difference-in-differences
$\bigl(\mathrm{MD}-\mathrm{TXT}\bigr)_{\mathrm{CC}}-
 \bigl(\mathrm{MD}-\mathrm{TXT}\bigr)_{\mathrm{Codex}}$
and a 95\% nonparametric bootstrap interval using 20{,}000 resamples of
the ten runs in each cell. These intervals are unadjusted for six comparisons
and exploratory given the small per-cell sample. We define a narrow final numerical outcome before examining
traces: application revenue for Revenue, and application revenue, total cost,
and productivity for Productivity. Accordingly, ``applicable'' refers to the
skill-specific set of one or three outcome checks; no checks are dropped because
of execution errors. Other checks are used to measure additional
workflow deviations; this grouping does not imply equal operational severity.
As a secondary decomposition, the broader final-state layer comprises seven
checks per skill: Revenue's application and platform revenue, application and
group call counts, both normalization sums, and response structure; and
Productivity's revenue, cost, productivity, EC2 and Kubernetes resource shares,
the normalization sum, and response structure. All remaining checks form the
trajectory layer.
Cost is the per-run USD estimate recorded by each harness. We summarize it
descriptively because provider pricing, cache accounting, and model contracts
are not controlled experimental factors.

\section{Results}

\subsection{Cross-Model and Cross-Harness Results}

Table~\ref{tab:auto} reports average suite pass rates across all
240~automatic trials. Revenue portfolio deviations were at most one cent,
yet no configuration reached a 100\% mean suite score. Because the score
combines outcome and process checks, we use trace-level diagnostics---rather
than the aggregate pass rate alone---to identify process deviations.

\begin{table}[ht]
\centering
\caption{Automatic evaluation: mean pass rate (\%) $\pm$ standard deviation
across 10~trials per cell. Two skills (Revenue: 104~test cases;
Productivity: 80~test cases), two harnesses, three models, two specification variants.
Gap = MD$\,-\,$TXT (pp); positive = MD scores higher.
Robustness: R ($|$gap$|<2$\,pp), M ($2$--$8$\,pp), S ($>8$\,pp).}
\label{tab:auto}
\resizebox{\columnwidth}{!}{%
\begin{tabular}{llccccc}
\toprule
\textbf{Skill} & \textbf{Harness} & \textbf{Model} &
  \textbf{MD (\%)} & \textbf{TXT (\%)} & \textbf{Gap} & \textbf{Rob.}\\
\midrule
\multirow{6}{*}{Revenue}
  & \multirow{3}{*}{Claude Code}
    & GPT-5.6-Sol & $94.5\!\pm\!1.0$ & $95.7\!\pm\!4.6$ & $-$1.2 & R \\
  && Opus~4.8    & $96.5\!\pm\!0.5$ & $86.8\!\pm\!2.3$ & $+$9.7 & S \\
  && Sonnet~4.6  & $95.0\!\pm\!2.4$ & $88.8\!\pm\!5.4$ & $+$6.3 & M \\
\cmidrule{2-7}
  & \multirow{3}{*}{Codex}
    & GPT-5.6-Sol & $95.0\!\pm\!2.2$ & $82.1\!\pm\!6.4$ & $+$12.9 & S \\
  && Opus~4.8    & $95.5\!\pm\!1.1$ & $83.7\!\pm\!2.3$ & $+$11.8 & S \\
  && Sonnet~4.6  & $92.2\!\pm\!2.9$ & $92.6\!\pm\!4.2$ & $-$0.4  & R \\
\midrule
\multirow{6}{*}{Prod.}
  & \multirow{3}{*}{Claude Code}
    & GPT-5.6-Sol & $99.6\!\pm\!0.8$ & $95.0\!\pm\!0.8$ & $+$4.6 & M \\
  && Opus~4.8    & $98.0\!\pm\!0.6$ & $95.9\!\pm\!1.3$ & $+$2.1 & M \\
  && Sonnet~4.6  & $98.3\!\pm\!1.5$ & $95.6\!\pm\!1.5$ & $+$2.6 & M \\
\cmidrule{2-7}
  & \multirow{3}{*}{Codex}
    & GPT-5.6-Sol & $90.9\!\pm\!6.3$ & $91.9\!\pm\!4.0$ & $-$1.0 & R \\
  && Opus~4.8    & $98.0\!\pm\!1.6$ & $95.3\!\pm\!3.6$ & $+$2.8 & M \\
  && Sonnet~4.6  & $92.3\!\pm\!4.9$ & $91.6\!\pm\!8.1$ & $+$0.6 & R \\
\bottomrule
\end{tabular}
}
\end{table}

\paragraph{Process checks add detection beyond final numerical outcomes.}
Separating the narrow final numerical outcome defined in \S6 from all
additional checks, 175 of 240 trials pass
every applicable final numerical check, 162 of those 175 (92.6\%; Wilson
95\% CI: 87.7--95.6\%) have at least one additional evaluator-detected
deviation. All 120 Revenue trials pass the final revenue check, and all 120
have at least one additional evaluator-detected deviation. Thus, final-value
checks alone would miss deviations in 67.5\% of all trials.

The conclusion is robust to two error filters. Of the 33 runs with external
errors, 31 pass all applicable final numerical checks and all 31 retain an
additional deviation. Excluding those runs gives 131/144 (91.0\%; Wilson 95\%
CI: 85.2--94.6\%). Separately, excluding every run with an unrecovered error
gives 136/147 (92.5\%; Wilson 95\% CI: 87.1--95.8\%).
Across the complete suite, 227 of 240 runs (94.6\%; Wilson 95\%
CI: 91.0--96.8\%) fail at least one check.

Among the 162 outcome-correct but flagged trials, non-exclusive root families
are data or numerical consistency (106), missing phase or required tool (66),
allowlist violations (55), redundancy or call-count violations (43),
scope/filter violations (13), response format (2), and tool execution (1).
These counts distinguish the observed gap from a single evaluator artifact,
but they do not assign severity and can sum beyond 162.
Because many allowlist hits are harness utilities or schema operations, we also
remove the \texttt{no\_irrelevant\_tools} check entirely as a sensitivity test;
all 162 outcome-correct flagged runs still retain another failed check.

\paragraph{Trajectory checks add detection beyond a broader final state.}
Of 240 trials, 164 pass all seven skill-specific final-state checks defined in
\S6. Among these, 151 (92.1\%; Wilson 95\% CI: 86.9--95.3\%) still violate at
least one trajectory check; removing \texttt{no\_irrelevant\_tools} leaves this
count unchanged. The skill-level rates are 109/109 for Revenue and 42/55
(76.4\%) for Productivity. This secondary decomposition shows that the
increment is not solely an artifact of comparing a large suite with one or
three numerical values. It remains an authored-invariant analysis: a flagged
trajectory is not necessarily a confirmed operational defect.

\paragraph{Dependency attribution descriptively compresses failures.}
A run contains a median of 4 failed checks (mean 6.34), but only 2 root
failures (mean 2.65), a 58\% reduction in items requiring initial diagnosis.
This is graph-based descriptive compression; root labels were not independently
validated as causal explanations.
For Revenue, the leading roots are incorrect staged-utilization arguments
(75/120), use of a tool outside the evaluator allowlist (42/120), and missing
database readback (24/120). For Productivity, they are incorrect application
inputs to the computation tool (73/120), incorrect total cost (40/120),
application-filtered Kubernetes queries (38/120), and application-filtered EC2
queries (31/120). These are trial incidences, not mutually exclusive events.

\paragraph{Harness-dependent specification sensitivity.}
Revenue Apportionment reveals the sharpest MD--TXT divergence.
On Claude~Code, GPT-5.6-Sol is variant-robust ($-$1.2\,pp) while Opus~4.8
and Sonnet~4.6 are sensitive or moderately sensitive ($+$9.7 and
$+$6.3\,pp respectively).
On Codex, the pattern reverses: Sonnet~4.6 becomes the most robust model
($-$0.4\,pp) while GPT-5.6-Sol and Opus~4.8 exhibit the largest TXT
degradation in the study ($+$12.9 and $+$11.8\,pp).
Productivity Apportionment shows smaller MD--TXT gaps: all
models on Claude~Code are moderately sensitive (2.1--4.6\,pp), while
Sonnet~4.6 and GPT-5.6-Sol on Codex are robust ($<$2\,pp).
Bootstrap difference-in-differences intervals support a harness-dependent
variant effect for Revenue GPT ($-14.0$ pp, 95\% CI $[-18.5,-8.9]$),
Revenue Opus ($-2.1$ pp, $[-4.2,-0.2]$), Revenue Sonnet
($+6.6$ pp, $[2.0,11.2]$), and Productivity GPT
($+5.6$ pp, $[1.3,10.0]$). Productivity Opus
($-0.6$ pp, $[-3.1,1.8]$) and Sonnet ($+2.0$ pp, $[-4.3,6.9]$) are
inconclusive because their intervals include zero. These results show configuration-level
sensitivity, but the paired files do not isolate formatting from content.
A team migrating harnesses should therefore re-evaluate each complete skill
specification rather than assuming an MD or TXT effect transfers.

\paragraph{Harness divergence.}
Beyond specification sensitivity, Claude~Code consistently achieves higher
TXT-variant accuracy on Revenue Apportionment across all models
(CC~TXT: 86.8--95.7\% vs.\ Codex~TXT: 82.1--92.6\%).
Productivity Apportionment shows greater parity between harnesses for MD,
but Codex yields lower absolute accuracy for GPT-5.6-Sol
(90.9\% vs.\ 99.6\%), indicating that harness behavior
interacts with skill complexity.
However, unrecovered errors are concentrated in Codex runs (38 versus 5 for
Claude Code across the study; 11 versus 4 within Revenue), and all 20
Productivity Codex--Sonnet runs contain one. These contrasts may therefore
partly reflect execution reliability rather than instruction following;
detailed cell-level clean-run counts appear in Appendix~A. Because no clean
Productivity Codex--Sonnet run remains, clean-execution effects for that
configuration are not estimable from these data.

\paragraph{Recorded execution cost.}
Harness telemetry reports \$859.34 across 240 runs (mean \$3.58, median
\$1.23). These recorder-estimated observational costs are not causal harness
comparisons; Appendix~B reports runtime, cache accounting, and all 24 cells.

\section{Discussion}

\subsection{Template Test Cases as Reusable Regression Artifacts}

Template test cases with runtime-resolved placeholders function as reusable
regression artifacts for enterprise agentic skills, analogous to unit tests in
software CI/CD.
Each time the skill specification is revised, a new model version is
deployed, or a tool API changes, the same template test cases
re-execute against the new version and immediately surface any behavioral
regressions at the process level.
This reuse is possible precisely because placeholders are resolved at
evaluation time from live API state, not hardcoded at test-writing time:
the test that checks ``database readback was called before allocation''
is as meaningful for the v1.0 specification as it is for v1.7, and
for Sonnet~4.6 as it is for Opus~4.8 or GPT-5.6-Sol.
After removing six legacy checks retained in one archived result, all 24
model--harness--variant--skill conditions use one registered suite per skill.
Across archived materializations, the same check IDs cover four distinct Revenue
and 16 distinct Productivity expected-value states, demonstrating runtime
resolution under observed reference variation. This does not test controlled
input or API evolution or establish maintenance-free validity after semantic
changes; Appendix~A reports a post hoc staleness analysis. Prospectively, the
report could serve as a merge gate by flagging new roots or a configurable
root-count threshold, but we do not evaluate production merge decisions or
repair effectiveness.

\subsection{Evaluation Surfaces Specification Gaps}

Evaluator flags exposed implicit policy choices in EC2 memory and cost
allocation, DB readback ordering, and EC2 call scope. Continuous evaluation can
therefore act as a \emph{specification completeness tool} as well as a regression
detector. We reserve \emph{violation} for a trace that breaks an authored
invariant; domain review must distinguish confirmed defects, benign alternatives,
and specification ambiguities. Appendix~A details these cases.

\subsection{Specification-Revision Sensitivity Across Harnesses}

A key finding from the 240-trial automatic evaluation is that sensitivity to
the paired MD/TXT specification revisions can depend jointly on model,
harness, and skill type.
GPT-5.6-Sol is variant-robust on Claude~Code for Revenue Apportionment
($-$1.2\,pp gap) yet becomes the most variant-sensitive model on Codex
($+$12.9\,pp).
Conversely, Sonnet~4.6 is moderately sensitive on Claude~Code
($+$6.3\,pp) but variant-robust on Codex ($-$0.4\,pp).
Bootstrap interaction intervals exclude zero for all three Revenue
comparisons and for GPT on Productivity; the Productivity Opus and Sonnet
comparisons remain inconclusive. The supported inversions show that the
effect of a complete specification revision can change across harnesses.
Because MD and TXT also differ in content, the study cannot attribute these
effects to formatting itself. A specification validated on one harness
should therefore be re-evaluated after migration.

\subsection{Outcome Accuracy Does Not Establish Process Conformance}

Computation tools are the system's operational baseline in every reported
condition; the experiment therefore estimates neither a tool-encapsulation
effect nor arithmetic instability without tools. Its supported claim is
observational: even among 175 runs with correct final numerical outcomes, 162
contain another evaluator-detected deviation. More conservatively, 151/164
runs that pass all seven broader final-state checks still violate a trajectory
check. These deviations span multiple root families and persist when the
allowlist check is removed. Process-level regression tests therefore add
coverage beyond the chosen final state, while the operational severity of each
deviation remains domain specific.

\section{Conclusion}

Across 240 trials, 162/175 runs with correct final numerical outcomes contained
another evaluator-detected deviation; under the broader final-state baseline,
151/164 passing runs still violated a trajectory check. Dependency attribution
reduced 6.34 failed checks to 2.65 roots per run on average. These results show
that process-level regression tests complement final-state evaluation, while
template reuse and harness-dependent sensitivity motivate re-evaluation as
enterprise skills evolve.

\bibliographystyle{plainnat}
\bibliography{references}

\appendix
\section{Limitations}

This study evaluates one enterprise system and two related skills.
Productivity extends Revenue with external cost sources, cost allocation,
NULL handling, and a different persistence path; generalization to other
skill types, tool suites, or non-quantitative outputs remains open.

The 10-trial-per-cell sample supports conclusions about central tendency
and relative rankings but limits statistical power for small effect
sizes. Bootstrap intervals are exploratory; two of the six Productivity
model comparisons do not establish a specification-by-harness interaction.

MD and TXT are bundled specification variants rather than a controlled
format intervention. They differ in syntax, length, detail, and sometimes
semantics---most notably, Productivity TXT adds exclusion, normalization,
combined-cost-query, and CSV-export requirements. The observed gaps cannot
therefore be causally attributed to file format or condensation alone.

The aggregate suite score combines heterogeneous outcome and process
checks. It does not encode failure severity, and dependency-linked checks
are not statistically independent; category-specific root-failure rates
are therefore needed before interpreting small score differences
operationally.

The traces contain execution noise: 83/240 runs record at least one tool or
harness error. External and unrecovered errors are overlapping rather than
nested categories: 175 runs have neither, 32 have an unrecovered but no
external error, 22 have an external but no unrecovered error, and 11 have
both. Thus 33 runs have external errors and 43 have unrecovered errors.
External-error totals are nearly balanced by harness (Claude Code: 16;
Codex: 17) but uneven across cells, whereas unrecovered errors are concentrated
in Codex runs (Claude Code: 5; Codex: 38). Harness and variant contrasts may
therefore partly reflect infrastructure and execution reliability. We retain
all runs as an intention-to-evaluate analysis; future work should preregister
retry and exclusion rules and report cell-level clean-run contrasts.
The aggregate unrecovered-error sensitivity is not a substitute for such
cell-level analysis: excluding these runs removes all 20 Productivity
Codex--Sonnet observations, so it cannot validate that configuration or its
interaction estimate.

\begin{table}[ht]
\centering
\caption{Clean runs per 10-trial cell. A run is clean when it has neither an
external nor an unrecovered error. Zero-clean cells are not estimable under a
clean-execution analysis.}
\label{tab:clean-runs}
\small
\setlength{\tabcolsep}{4pt}
\begin{tabular}{lllrr}
\toprule
\textbf{Skill} & \textbf{Harness} & \textbf{Model} & \textbf{MD} & \textbf{TXT} \\
\midrule
\multirow{6}{*}{Revenue}
 & \multirow{3}{*}{Claude Code} & GPT-5.6-Sol & 6 & 9 \\
 && Opus~4.8 & 10 & 4 \\
 && Sonnet~4.6 & 10 & 2 \\
 & \multirow{3}{*}{Codex} & GPT-5.6-Sol & 9 & 9 \\
 && Opus~4.8 & 10 & 8 \\
 && Sonnet~4.6 & 1 & 5 \\
\midrule
\multirow{6}{*}{Productivity}
 & \multirow{3}{*}{Claude Code} & GPT-5.6-Sol & 9 & 10 \\
 && Opus~4.8 & 10 & 10 \\
 && Sonnet~4.6 & 10 & 10 \\
 & \multirow{3}{*}{Codex} & GPT-5.6-Sol & 9 & 10 \\
 && Opus~4.8 & 9 & 5 \\
 && Sonnet~4.6 & 0 & 0 \\
\bottomrule
\end{tabular}
\end{table}

The frequent EC2-memory disagreement was not introduced by a specification
revision. The tool can return host/JVM memory where no defensible
per-application split exists, whereas the authored invariant expects zero.
Writing the test exposed an implicit policy choice: \emph{what should be stored
when a per-application split is impossible?} The raw results therefore identify
a specification gap, but do not by themselves prove that zero is the uniquely
correct value. The same issue arises when the Productivity oracle expects zero
EC2 cost but some executions allocate nonzero Cloudability cost. This pattern
recurred for DB readback ordering and EC2 call scope. In each case, specifying
the expected behavior surfaced an assumption that the skill specification had
not made explicit. This distinction permits multiple valid trajectories rather
than equating one canonical trace with operational correctness.

The evaluator is normative: in 71/240 trials its ``irrelevant tool'' check is a
root failure, sometimes for harness utilities, schema initialization, or
namespace aliases of unestablished harm. Ordering and zero-valued EC2
memory/cost checks also encode workflow policy. The unchanged 162/175 result
after removing the allowlist check shows that this issue does not explain the
headline gap, but domain review, equivalent-trace tests, and severity labels
are still needed to separate harmful from benign behavior.

Live-API ground truth can drift between execution and deferred evaluation,
so immutable response snapshots would strengthen reference integrity. Our
targeted two-author audit found agreement on 57 of 60 randomly sampled
LLM-judged cases (95\%), but this should not be interpreted as general judge
accuracy: the sample was drawn over judgments rather than runs, its coverage
across check types was not stratified, the reviewers were authors rather than
external annotators, and the audit does not separately characterize the
failure-only rescue path. Calculator-backed arithmetic reduces but does not
remove semantic-judgment risk. Moreover, the judge model (Claude Sonnet 4.6)
is also one of the evaluated models, so correlated or self-preference bias may
affect judgments of its traces.
The manually authored invariants and dependency graph may require maintenance
as tool semantics evolve. The experiments compare fixed configurations rather
than a longitudinal API or model upgrade, so persistence under actual evolution
remains untested. Both skills also use one portfolio, one revenue value, and one
date window each; repeated stochastic trials do not provide environmental input
diversity. In a post hoc staleness simulation, freezing the modal Productivity
expected values would place at least one value outside its registered tolerance
in 80/120 Productivity runs, whereas observed Revenue variation remained within
tolerance. This counterfactual and the materialization analysis support runtime
resolution under observed reference variation, not controlled evolution.
Finally, the framework diagnoses but does not repair failures.

The cost analysis uses harness-recorded USD estimates rather than reconciled
provider invoices. Pricing rules, cache accounting, and provider contracts can
differ across harnesses and models, and unusually expensive retries can create
heavy tails. Consequently, the cost results characterize these 240 executions
and should not be interpreted as an intrinsic price advantage of a harness.

\section{Cost and Runtime Results}

Across all 240 runs, harness telemetry reports \$859.34 total cost: mean
\$3.58 (SD \$4.50), median \$1.23, and range \$0.45--\$30.23 per run.
Claude Code averages \$1.04 versus \$6.12 for Codex under the balanced design.
Within each skill and pooling variants, Codex/Claude-Code mean-cost ratios are
5.9--6.0$\times$ for Opus and 8.8--10.2$\times$ for Sonnet, but 1.0--1.4$\times$
for GPT. For Claude models, cache-read tokens comprise 90.7--92.9\% of recorded
Claude Code tokens versus 0\% on Codex, consistent with caching as a major
accounting difference. These are recorder-estimated, observational costs rather
than invoice-validated causal effects. Mean runtime in seconds for Claude
Code/Codex is 416/238 for GPT, 254/450 for Opus, and 284/344 for Sonnet;
the Codex--Opus mean is inflated by one 7{,}152-second run (median: 241 seconds).

Table~\ref{tab:cost-runtime} aggregates the balanced design over both skills
and both specification variants ($n=40$ per harness--model cell). Runtime is
reported alongside cost because retries and long-running failures can affect
both measures.

\begin{table}[ht]
\centering
\caption{Recorded cost, runtime, and cache-read share, aggregated over both
skills and variants ($n=40$ per row). Runtime is mean/median seconds.}
\label{tab:cost-runtime}
\small
\setlength{\tabcolsep}{5pt}
\begin{tabular}{llrrr}
\toprule
\textbf{Harness} & \textbf{Model} & \textbf{Cost/run} & \textbf{Runtime} & \textbf{Cache} \\
\midrule
\multirow{3}{*}{Claude Code}
 & GPT-5.6-Sol & \$0.68 & 416/350 & 0.0\% \\
 & Opus~4.8    & \$1.54 & 254/212 & 92.7\% \\
 & Sonnet~4.6  & \$0.91 & 284/259 & 91.2\% \\
\midrule
\multirow{3}{*}{Codex}
 & GPT-5.6-Sol & \$0.75 & 238/162 & 45.0\% \\
 & Opus~4.8    & \$9.11 & 450/241 & 0.0\% \\
 & Sonnet~4.6  & \$8.49 & 344/320 & 0.0\% \\
\bottomrule
\end{tabular}
\end{table}

Table~\ref{tab:cost} reports every cell in the current 240-run design. Each
entry is the mean recorded USD cost per run $\pm$ sample standard deviation
over 10 trials; no run has missing or zero cost telemetry.

\begin{table}[ht]
\centering
\caption{Harness-recorded cost per run in USD (mean $\pm$ SD; $n=10$ per cell).}
\label{tab:cost}
\small
\setlength{\tabcolsep}{5pt}
\begin{tabular}{llrrr}
\toprule
\textbf{Skill} & \textbf{Harness} & \textbf{Model} & \textbf{MD} & \textbf{TXT} \\
\midrule
\multirow{6}{*}{Revenue}
 & \multirow{3}{*}{Claude Code} & GPT-5.6-Sol & $0.99\!\pm\!0.27$ & $0.78\!\pm\!0.55$ \\
 & & Opus~4.8   & $1.96\!\pm\!0.68$ & $1.64\!\pm\!0.19$ \\
 & & Sonnet~4.6 & $1.03\!\pm\!0.23$ & $1.26\!\pm\!0.29$ \\
\cmidrule{2-5}
 & \multirow{3}{*}{Codex} & GPT-5.6-Sol & $0.71\!\pm\!0.19$ & $1.00\!\pm\!0.14$ \\
 & & Opus~4.8   & $7.25\!\pm\!1.42$ & $13.87\!\pm\!6.44$ \\
 & & Sonnet~4.6 & $7.63\!\pm\!0.73$ & $12.47\!\pm\!5.50$ \\
\midrule
\multirow{6}{*}{Productivity}
 & \multirow{3}{*}{Claude Code} & GPT-5.6-Sol & $0.48\!\pm\!0.01$ & $0.46\!\pm\!0.01$ \\
 & & Opus~4.8   & $1.28\!\pm\!0.40$ & $1.25\!\pm\!0.23$ \\
 & & Sonnet~4.6 & $0.64\!\pm\!0.18$ & $0.72\!\pm\!0.13$ \\
\cmidrule{2-5}
 & \multirow{3}{*}{Codex} & GPT-5.6-Sol & $0.61\!\pm\!0.12$ & $0.70\!\pm\!0.09$ \\
 & & Opus~4.8   & $7.84\!\pm\!5.55$ & $7.49\!\pm\!1.37$ \\
 & & Sonnet~4.6 & $6.57\!\pm\!0.38$ & $7.30\!\pm\!1.85$ \\
\bottomrule
\end{tabular}
\end{table}

\end{document}